\pdfoutput=1

\documentclass[11pt]{article}

\usepackage{acl}

\usepackage{times}
\usepackage{latexsym}
\usepackage{booktabs}

\usepackage{algorithm}
\usepackage{multirow}
\usepackage[noend]{algorithmic}
\usepackage{wrapfig}
\usepackage{framed}

\definecolor{blue}{HTML}{4285f4}
\definecolor{LightGrey}{HTML}{d9d9d9}

\usepackage[T1]{fontenc}

\usepackage[utf8]{inputenc}

\usepackage{microtype}

\usepackage{inconsolata}

\usepackage{amsmath}
\usepackage{graphicx}

\usepackage{multirow}
\definecolor{green}{HTML}{34a853}
\definecolor{lightgreen}{HTML}{d9ead3} 
\definecolor{seagreen}{HTML}{3CB371} 
\definecolor{darkgreen2}{HTML}{38761d}
\definecolor{lightgreen2}{HTML}{b6d7a8}

\definecolor{redberry}{HTML}{cc4125} 
\definecolor{lightredberry}{HTML}{cc4125} 
\definecolor{lightred}{HTML}{e06666}
\definecolor{darkpurple1}{HTML}{674ea7}

\definecolor{purple}{HTML}{9900ff} 
\definecolor{lightpurple}{HTML}{b4a7d6} 
\definecolor{lightpurple1}{HTML}{8e7cc3}

\definecolor{gray}{HTML}{cccccc}
\definecolor{lightgray1}{HTML}{d9d9d9}
\definecolor{lightgray2}{HTML}{efefef}
\definecolor{darkgray4}{HTML}{434343}

\usepackage{graphicx}
\graphicspath{ {./images/} }

\definecolor{blue}{HTML}{4285f4} 
\definecolor{darkblue}{HTML}{0b5394} 
\definecolor{lightblue}{HTML}{9fc5e8} 
\definecolor{lightblue3}{HTML}{cfe2f3} 
\definecolor{lightcornflowerblue2}{HTML}{a4c2f4} 
\definecolor{lightcornflowerblue3}{HTML}{c9daf8} 
\definecolor{darkcornflowerblue3}{HTML}{1c4587} 

\definecolor{orange}{HTML}{ff9900} 
\definecolor{lightorange}{HTML}{f9cb9c} 
\definecolor{lightorange3}{HTML}{fce5cd} 
\definecolor{darkorange}{HTML}{FF8C00} 
\definecolor{darkorange1}{HTML}{e69138} 

\definecolor{lightyellow2}{HTML}{ffe599}
\definecolor{lightyellow3}{HTML}{fff2cc}

\usepackage{enumitem}
\usepackage{booktabs}
\usepackage{times}
\usepackage{latexsym}
\title{Crafting In-context Examples according to LMs' Parametric Knowledge}
\usepackage{subcaption}
\usepackage{bbm}

\def\star{\textsuperscript{*}}
\newcommand{\blu}[1]{\textcolor{blue}{#1}}
\newcommand{\red}[1]{\textcolor{red}{#1}}

\author{Yoonsang Lee$^{\diamondsuit,\heartsuit}$\thanks{$^{*}$Equal Contribution, work was done at UT Austin. }~~   Pranav Atreya$^{\spadesuit,{*}}$ ~~ Xi Ye$^{\diamondsuit}$ ~~ Eunsol Choi$^{\diamondsuit}$ \\
$^\diamondsuit$The University of Texas at Austin, $^\heartsuit$Seoul National University, $^\spadesuit$UC Berkeley \\
 \hspace{0.5em} {\texttt{lysianthus@snu.ac.kr, pranavatreya@berkeley.edu}} \\
  {\texttt{xiye@cs.utexas.edu, eunsol@utexas.edu}}
 }

\begin{document}
\maketitle
\begin{abstract}
In-context learning can improve the performances of knowledge-rich tasks such as question answering. 
In such scenarios, in-context examples trigger a language model (LM) to surface information stored in its parametric knowledge. We study how to better construct in-context example sets, based on whether the model is aware of the in-context examples. We identify `known' examples, where models can correctly answer from their parametric knowledge, and `unknown' ones. Our experiments show that prompting with `unknown' examples decreases the performance, potentially as it encourages hallucination rather than searching for its parametric knowledge. Constructing an in-context example set that presents both known and unknown information performs the best across diverse settings. We perform analysis on three multi-answer question answering datasets, which allows us to further study answer set ordering strategies based on the LM's knowledge of each answer. Together, our study sheds light on how to best construct in-context example sets for knowledge-rich tasks.\footnote{Our code is available at \url{https://github.com/lilys012/known_examples}.} 
\end{abstract}
\section{Introduction}

\begin{figure}[tb]
    \centering
    \includegraphics[width=0.5\textwidth]{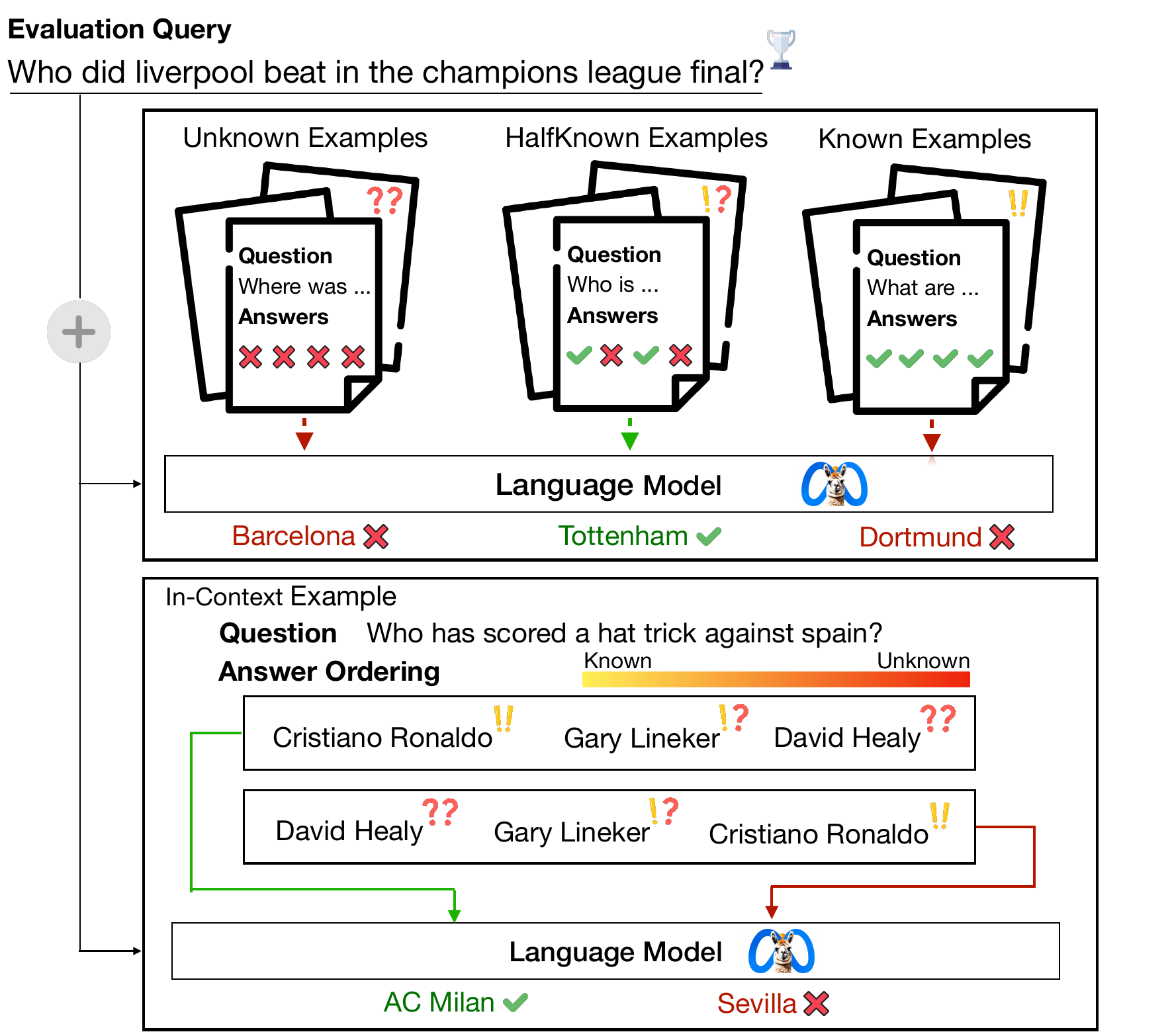}    
    \caption{We study how an LM's knowledge of in-context examples impacts its effectiveness. On the top box, we construct three sets of in-context examples, {Unknown}, {HalfKnown}, and {Known}, differing in their difficulty 
    (Section \ref{pkexp}). On the bottom box, we construct two in-context examples, which contain the same question and answer set, but the answers are sorted differently: one in increasing amount of parametric knowledge and one in reverse (Section \ref{sec:ansordering}, \ref{sec:ordering_results}).}
    \label{fig:overview}
\end{figure}

Large language models (LLMs) can perform competitively on knowledge-rich tasks such as question answering via in-context demonstrations~\citep{brown2020language}. In such scenarios, in-context examples are used not only to teach the LLM the mapping from inputs to outputs, but also to invoke the LLM's parametric knowledge~\citep{liu2021makes, Agrawal2022IncontextES}. Given such role of in-context examples, we examine how the LLM's parametric knowledge of in-context examples impact the effectiveness of in-context examples. 

Let's imagine a very challenging in-context example set, where LLMs cannot answer any of in-context examples from its parametric knowledge. For example, in-context examples can query knowledge about recent events that happened after pre-training~\citep{Lazaridou2021MindTG}. These in-context examples may teach the model to generate plausible-looking responses, but also encourage hallucination as a result. On the other hand, if we only provide in-context examples where LLM can easily answer, would LLM learn to make an educated guess on more challenging evaluation examples? 

We pose a suite of research questions connecting parametric knowledge of an LM on in-context examples and its impact on model predictions. Figure~\ref{fig:overview} provides our study overview. We mainly focus on multi-answer QA datasets~\cite{min2020ambigqa,malaviya2023quest,amouyal2022qampari} since they allow selecting and ordering the answers among multiple answers. Furthermore, we evaluate on a math QA dataset~\cite{Cobbe2021TrainingVT} and two NLI datasets~\cite{Dagan2005ThePR, Bowman2015ALA}, which require reasoning from LLMs.

We first compare providing `known' or `unknown' in-context examples (Section~\ref{pkexp}). We operationalize `known' in-context examples as those LM can correctly predict with in-context learning. We do not observe a clear winner between two choices, with results varying depending on the dataset. Throughout all datasets, however, providing a set of in-context examples that have a mixture of known and unknown information leads to superior performance compared to a set consisting solely of known or unknown in-context examples. 

Our next analysis focuses on the ordering of multi-answer set while fixing in-context example set (Section~\ref{sec:ansordering}, \ref{sec:ordering_results}). Compared to randomly ordering valid answers, semantically meaningful ordering brings substantial changes in model predictions. Even alphabetical ordering of answer set changes predicted answers substantially, prompting the model to generate 1.5 more answers on average than when shown a randomly sorted answer set. We further find that ordering the answer set of in-context examples in descending order of model knowledge often leads to performance gains. Together, our work suggests best practices for crafting in-context examples, with relation to their parametric knowledge, for knowledge-intensive tasks.
\section{Experimental Settings}
We first describe our evaluation setting which centers around multi-answer QA datasets. 

\subsection{Dataset}
We evaluate on three multi-answer QA datasets: (1) AmbigQA~\citep{min2020ambigqa} contains a subset of questions from the Natural Questions~\citep{kwiatkowski2019natural} dataset, namely those marked as ambiguous in the sense that depending on the interpretation, they can have multiple correct answers. 
(2) QAMPARI~\citep{amouyal2022qampari} consists of questions whose set of correct answers necessarily span multiple paragraphs in the document from which they were retrieved. 
The dataset was originally developed to evaluate retrieval methods, and we repurpose it to create a challenging closed-book QA setting. 
(3) QUEST~\citep{malaviya2023quest} dataset is constructed by formulating queries that define implicit set operations over Wikipedia entities. We report the dataset statistics in Appendix~\ref{appendix:dataset}.

\subsection{Evaluation Metrics}

Given a question $q$, the model will predict a set of answers $\hat{a} = \{a_1, a_2, ..., a_m\}$, where each $a_i = (w_{i_1}, w_{i_2}, ..., w_{i_{|a_i|}})$ is a sequence of tokens for a single answer. We denote $a^* = \{a^*_1, a^*_2, ..., a^*_n\}$ as the ground truth answers to the same question. 

We use standard token match metrics for evaluating answer accuracy, Exact Match (EM) and F1-score \cite{Joshi2017TriviaQAAL}. EM assigns a score of $1$ if the predicted answer equals the ground truth answer, while F1-score is calculated over the tokens in the answer.
We use metrics for multi-answers introduced in prior work~\cite{min2020ambigqa}, which we describe below for completeness.

\paragraph{Answer-level Exact Match ($\mathbf{F1_{EM}}$)}
As predicting the exact ground truth answer set correctly is very challenging, we report the F1-score of answer-level exact match, denoted as $F1_{EM}$. For an answer $a$ and reference answers set $S$, we define a correctness score $c(f, a, S) = f(a, S)$ with respect to function $f$. We use $f(a, S) = \mathbbm{1}(a \in S)$ here. Then, we calculate the F1-score over set-level precision and recall according to $c$.
\[P = \frac{\sum_{i=1}^{m} c(f, a_i, a^*)}{m}, R = \frac{\sum_{j=1}^{n} c(f, a^*_j, \hat{a})}{n}\]
\[F1_{EM} = \frac{2 \times P \times  R}{P + R}\]

\paragraph{Answer-level F1 ($\mathbf{F1_{F1}}$)} The generated answer may be semantically equivalent to one of the ground truth answers, without being lexically equivalent (e.g., "Friends" and "The TV show Friends").
To account for such semantic equivalences, we use $F1$ score between the tokens of two answer strings instead of the exact match as a correctness score, $f(a, S) = max_{a' \in S}(F1(a, a'))$. Then, we compute F1-score over set-level precision and recall as above. 

\begin{table*}
\footnotesize
\begin{center}
\begin{tabular}{l|cc|cc|cc}
\toprule
& \multicolumn{2}{c|}{AmbigQA\textsubscript{dev}} & \multicolumn{2}{c|}{QAMPARI\textsubscript{dev}} & \multicolumn{2}{c}{QUEST\textsubscript{test}} \\
& Llama2 & GPT-3.5 & Llama2 & GPT-3.5& Llama2 & GPT-3.5\\
\midrule
Random & 18.0\:\:/ 28.9\: & 20.0\: / 31.6\: & 10.3\: / 20.8\: & 15.0\:\:/ 28.5\: & 3.4\: / 11.0\: & 6.0\:\:/ 16.6\: \\
Unknown & 17.2\star / 28.2\star & 20.3\star / 33.1\star & 10.9\star / 22.0\star & 14.8\:\:/ 27.9\star & 3.7\star / 11.9\star & 5.7\star / 15.8\star \\ 
HalfKnown & \textbf{18.5}\star / \textbf{29.5}\star & \textbf{21.6}\star / \textbf{33.2}\star & \textbf{11.3}\star / \textbf{22.6}\:\: & \textbf{15.5}\star / 28.2\star & \textbf{4.0}\star / 11.9\star & \textbf{6.3}\star / \textbf{17.4}\star \\
Known & 18.3\star / 29.0\star & 21.3\star / 33.1\star & 9.8\: / 19.7\: & 15.3\:\:/ \textbf{29.2}\star & 3.9\star / \textbf{12.0}\star & 5.4\star / 15.8\: \\
\bottomrule
\end{tabular}
\caption{
Results comparing known example and unknown example. We present $F1_{EM}$ and then $F1_{F1}$ in each cell. Using half-known example outperforms other settings. We put $^*$ on scores that are significantly different from that of Random in-context examples set, and bold the highest performing set for each metric.
}
\label{tab:answer_sets}
\end{center}
\end{table*}

\paragraph{Statistical Testing}
\label{lab:testing}
As our evaluation datasets are relatively small, we conduct paired bootstrap tests throughout most of our experiments, highlighting results that outperform baseline with p value of $\le 0.05$.
\subsection{Base Models} 
\paragraph{Language Model}
We evaluate on Llama2~\cite{touvron2023llama} (13B) language model mainly, and additionally OPT \cite{Zhang2022OPTOP} (13B) and GPT-3.5-turbo models to evaluate generalization.

\paragraph{In-Context Example Retriever}
Prior work~\cite{rubin2021learning} has established that using semantically similar in-context examples improves the performance of in-context learning significantly. Throughout our study, we often retrieve top 5 most similar in-context examples from the entire training set for each dataset to form the prompt. We place in-context examples in increasing order of similarity, such that the most similar example will be presented the closest to the evaluation question. We measure example similarities by encoding each {question} with a SimCSE model~\cite{gao2021simcse} and computing their dot product. 

\section{Known Examples vs. Unknown Examples}
\label{pkexp}

Prior work has studied a few characteristics of successful in-context example set, such as label distribution in the in-context example set \cite{min2022rethinking}. We evaluate in-context examples with respect to model's parametric knowledge, whether a `known' or `unknown' in-context example is better. We operationalize `known' ones as the ones where LLMs can get the answers correctly from its own parametric knowledge, and `unknown' ones as those that the model did not answer it correctly.

\subsection{In-Context Example Set Study}
\label{sec:known_set}
We create four sets of in-context examples, differing in its difficulty for a given LM.
\begin{itemize}[noitemsep,leftmargin=*]
    \item{\textbf{\textsc{Unknown}}}: examples for which the LM possesses no knowledge of the answers. Operationally, these are examples when LM is prompted with five most similar examples, LM will predict zero answers correctly (i.e. zero $\mathbf{F1_{EM}}$ score). 
    \item{\textbf{\textsc{Random}}}: randomly sampled examples. Since the LM possesses no knowledge to majority of the examples, these exhibit 0.18 $\mathbf{F1_{EM}}$ score on average. 
    \item{\textbf{\textsc{HalfKnown}}}: examples for which the LM possesses roughly half knowledge of the answers (i.e. 0.5 $\mathbf{F1_{EM}}$ score).
    \item{\textbf{\textsc{Known}}}: examples for which the LM possesses full knowledge of the answers (i.e. 1.0 $\mathbf{F1_{EM}}$ score). 
\end{itemize} 

As prior work~\cite{rubin2021learning} has established that the similarity of in-context example to the query correlates strongly with the model's performance, we control for this confounding factor. We compute the average similarity for each in-context example candidate to other in-context example candidates in the candidate set (training set). Then, we choose a fixed number of in-context examples whose average similarity value is close to the median value.\footnote{We choose 999 examples for AmbigQA and QAMPARI, and 499 for QUEST (as QUEST only has 1251 training examples), half from below median, half from above median. For QUEST, we could not find enough examples with where model score full $\mathbf{F1_{EM}}$ score, so we selected highest scoring examples. The mid-range is (0.245, 0.264), (0.294, 0.296), (0.326, 0.373) for AmbigQA, QAMPARI, and QUEST.}  From this candidate set, we sample five examples for each condition and use them as fixed in-context examples across all questions in the evaluation dataset. To further reduce randomness, we sample multiple sets of five example set for each condition and report the average performance (by default, four sets are sampled and two sets are sampled for \textsc{HalfKnown} and \textsc{Known} set in QUEST because of lack of examples with sufficient model knowledge). 

We present the performance of each in-context example set for three datasets with Llama2 and GPT-3.5 in Table \ref{tab:answer_sets}. We observe the \textsc{HalfKnown} in-context example set achieves strong performance consistently on both LMs. Since half-known examples contain both answers that the model knows and doesn't know, we hypothesize this may successfully prompt LMs to leverage parametric knowledge and to make educated guesses.

\begin{figure}[tbp]
    \centering
    \includegraphics[width=0.5\textwidth]{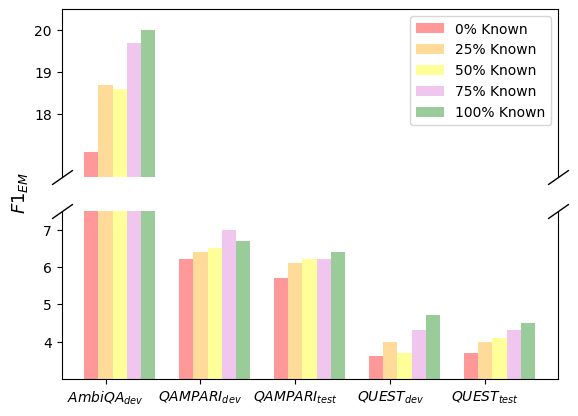}   
    \caption{
    Results of single answer study on Llama2 model.
    Only an answer at the $x$-th quantile of perplexities in decreasing order is presented in each in-context example.
    As the model gets exposed to more {known} answers, the performance tend to increase.
    }
    \label{fig:oneans}
\end{figure}

\subsection{Single Answer Study} \label{subsec:single_answer}
In this study, we further control for variability in the questions used in in-context examples. We fix the in-context example set and manipulate the multi-answer set, such that we provide only one answer from multi answer set for each in-context example. For example, if a question in in-context example is ``Who was the president of U.S.?'', we can either provide a famous president or a lesser-known president as an answer. Both are `correct' answers, but which answer would lead to better model performance? 

For each question in our evaluation set, we retrieve the top five most similar examples from the training set as in-context examples. We measure the perplexity of each answer to approximate how well LM `knows' the answer. For each example, a pair of question $q$ and gold answer set $\{a^*_1, a^*_2, ... a^*_n\}$, we form a prefix $p$ by prepending the top five most similar examples to the query $q$.\footnote{We present an example prefix in Appendix ~\ref{prompt_example}.
} Then, we compute the length normalized perplexity of each answer $a^*_i$ and prefix $p$ as follows: 
\begin{equation*} 
PP(a^*_i|p) = \prod^{|a^*_i|}_{j=1}{P(w_{ij}|p, w_{i1}, ..., w_{i(j-1)})}^{-\frac{1}{|a^*_i|}}
\end{equation*} We order the gold answer set in descending order of perplexity, and select an answer at the $x$-th quantile. This way, an answer at the 100\% quantile represents the most `known' answer, as its perplexity is the lowest among the gold answers.

Figure~\ref{fig:oneans} presents the $F1_{EM}$ score among various $x$-th quantile. We observe a clear trend across all three datasets, that using a `known' answer leads LM to generate more accurate answer. The overall F1 score is low, due to low recall, as these in-context examples are incomplete.\footnote{Since we provide only one answer for all in-context examples, LM predicts a single answer for each question.}
Nonetheless, this experiment affirms that crafting in-context examples by considering the model's parametric knowledge can impact the final performance.

\subsection{Extension to Other Tasks}
In this section, we explore the generalizability of our findings from Section~\ref{sec:known_set} to other tasks, single-answer QA and Natural Language Inference (NLI). 

\paragraph{Datasets}
For single-answer QA task, we select the GSM8K \cite{Cobbe2021TrainingVT} dataset, which is commonly used to assess the reasoning capabilities of LLMs.
For NLI task, we choose RTE \cite{Dagan2005ThePR} and SNLI \cite{Bowman2015ALA} datasets, two standard NLI benchmarks. For all datasets, we use their standard train, development, and test splits. 

We first evaluate each example in training set, to identify whether LM's parametric knowledge is sufficient to evaluate individual example correctly. We classify each example as correct, wrong, or invalid, where invalid indicates that the model did not produce an answer. For GSM8K, we use a fixed 8-shot example set taken from \citet{wei2022chain}. For NLI task, for each training example, we retrieve the top five most similar example from the training set (excluding itself) to form a 5-shot example set. We define the similarity as the dot product of two SimCSE embedding. After this process, each training example is labeled as correct, wrong or invalid. We throw away invalid examples and sample from correct and wrong set to form in-context example set, of varying degrees of difficulty.

\begin{table}[t]
\footnotesize
\begin{center}{
\scalebox{0.9}{
\begin{tabular}{r|cccc}
\toprule
&Unknown & Random &HalfKnown & Known \\ \midrule
GSM8K\textsubscript{test}&33.1& 34.8  & \textbf{36.4} & 32.0 \\ 
RTE\textsubscript{dev}	& 72.2&79.1&\textbf{79.8}&79.1 \\
SNLI\textsubscript{test}&62.7&69.3&\textbf{71.0}&68.6 \\
\bottomrule 
\end{tabular}
}}
\caption{Performance(accuracy) on GSM8K, RTE, and SNLI datasets. Accuracy is expressed as the percentage of correct answers over the entire test dataset.
}
\label{tab:extension}
\end{center}
\end{table}

\paragraph{In-Context Example Set}
 Unlike multi-answer QA, where examples can be partially correct, in these tasks, the examples are evaluated as either correct or incorrect. Therefore, we construct \textsc{HalfKnown} set by mixing easier and harder in-context examples as follows: 
\begin{itemize}[noitemsep,leftmargin=*]
    \item \textbf{\textsc{Unknown}} set includes randomly selected six examples that model answered incorrectly.
    \item \textbf{\textsc{Random}} set includes randomly selected six examples from entire training dataset. 
    \item \textbf{\textsc{HalfKnown}} set includes three correct and three wrong examples.
    \item \textbf{\textsc{Known}} set includes randomly selected six examples that model answered correctly.     
\end{itemize}

\paragraph{Result} We select six examples four times and report the averaged accuracy with Llama2 model in Table \ref{tab:extension}. On all three datasets, \textsc{HalfKnown} set achieves the highest accuracy, repeating the trend from multi-answer QA datasets. 
\section{Ordering Answers Based on LM's Knowledge}
\label{sec:ansordering}
Prior work suggests that the ordering of in-context examples significantly impacts the performance, with more relevant examples being most beneficial when placed last~\cite{Zhao2021CalibrateBU}. 
Yet, no prior work has studied how the ordering of answer set in in-context examples affects model generation and task performances.
We investigate this here. Following our previous study, our focus is on \textbf{parametric} knowledge of LMs being prompted. Specifically, we question whether placing answers based on how well the model knows about answers improves the performance. 

\label{sec:ordering_strategies}
We present strategies to order the answer set of each example, a pair of question $q$ and its gold answer set $a^*=\{a^*_1, a^*_2, ..., a^*_n\}$, which will be used as an in-context example.\footnote{As reordering process is computationally expensive, proportional to the number of answers, we only consider examples that have less than 20 answers. This results in exclusion of 1 example in AmbigQA, 8094 examples in QAMPARI, and none in QUEST.} We present two baselines and two methods (\textsc{Perplexity}, \textsc{Greedy}) for ordering the gold answer set of each in-context example based on model's parametric knowlege.

\paragraph{Baselines} The \textsc{random} baseline randomly orders answers and \textsc{alphabet} orders answers alphabetically. While alphabetical ordering is not relevant to model's parametric knowledge of the answer, prior work~\cite{Madaan2022ConditionalSG} has shown that consistent ordering of labels can improve the performance of fine-tuned LLM's predictions.

\paragraph{Knowledge-Aware Ordering}
We decide ordering based on the \textbf{perplexity} of individual answer given the prefix, or by performing \textbf{greedy} {constrained decoding}, given the prefix, on a smaller, restricted vocabulary set. We use the same prefix as in Section~\ref{subsec:single_answer}, a concatenation of five in-context examples. Each ordering strategy will yield two orderings of answers, which either sorts the answers in the descending order of model's parametric knowledge or ascending order (denoted as \textsc{Reverse}).

\begin{figure}
\begin{framed}
\small
\textbf{Input:} LM $\mathcal{M}$, Prefix $p$, Gold answer set $a^*=\{a^*_1, \dots, a^*_n\}$, where each gold answer is a token sequence (i.e., $a^*_i=(w_{i_1}, \dots w_{i_{|a^*_i|}})$) \\
\textbf{Output:} Ordered answer indices of the gold answer set\\
\begin{algorithmic}[1]
\STATE $I_1 \gets \{w_{1_1}, ..., w_{n_1}\}$
\STATE $u \gets 1$
\WHILE{$I_1 \neq \emptyset$}
    \STATE $t \gets 0$
    \REPEAT
        \STATE $t \gets t+1$
        \STATE $o_t \gets \textbf{argmax}_{w \in I_t} P_\mathcal{M}(w|p)$
        \STATE $p \leftarrow [p; o_t]$
        \STATE $I_{t+1} \gets \{w_{i_{t+1}} | w_{i_t} == o_t\}$
    \UNTIL{$\exists{a^*_{k_u}} == (o_1, \dots, o_t)$}
    \COMMENT{this assigns $k_u$ the index of completed answer}
    \STATE $I_1 \gets I_1 \setminus \{w_{{k_u}1}\}$
    \STATE $u \gets u+1$
\ENDWHILE
\STATE \textbf{return} $\{k_1, \dots, k_n\}$
\end{algorithmic}
\end{framed}
\caption{Algorithm for constrained decoding for \textsc{Greedy} ordering.
}\label{alg:grd}
\end{figure}

\begin{table*}
\footnotesize
\begin{center}
\begin{tabular}{llccccc}
\toprule
&&&&$S$\\
& & \textsc{Greedy} & \textsc{Reverse Greedy} & \textsc{Perplexity} & \textsc{Reverse Perplexity} &\textsc{Alphabet} \\ \midrule
&AmbigQA\textsubscript{dev} & 71.7 / 66.0 & 39.2 / 37.2 & 69.5 / 65.8 & 38.1 / 34.2 & 87.4 / 55.6 \\
&QAMPARI\textsubscript{dev} & 69.6 / 60.0 & 42.2 / 41.0 & 58.1 / 54.1 & 46.3 / 45.9 & 95.0 / 58.9 \\
$\mathcal{D}_e$&QAMPARI\textsubscript{test} & 70.0 / 65.7 & 43.0 / 41.7 & 58.8 / 55.8 & 45.0 / 44.2 & 94.9 / 58.1 \\
&QUEST\textsubscript{dev} & 78.4 / 63.9 & 47.2 / 45.8 & 57.1 / 51.5 & 49.3 / 48.5 & 95.7 / 52.1 \\
&QUEST\textsubscript{test} & 81.0 / 63.3 & 45.7 / 45.3 & 57.6 / 52.5 & 48.8 / 47.5 & 95.6 / 50.8 \\\midrule
\multicolumn{2}{c}{\textbf{Average}} & 74.1 & 43.5 & 60.2 & 45.5 & 93.7\\
\bottomrule
\end{tabular}
\caption{Percentage of generated answer ordering matching in-context examples answer ordering, where we use Llama2 for $\mathcal{M}$. In each cell, we present the percentage from using corresponding answer ordering strategy first  ($\phi(S, \mathcal{D}_t^S, \mathcal{D}_e,\mathcal{M})$) and the percentage for randomly ordering answers for control  ($\phi(S, \mathcal{D}_t^{S_\text{random}}, \mathcal{D}_e,\mathcal{M})$). 
}
\vspace{-5pt}
\label{tab:learning}
\end{center}
\end{table*}

\begin{itemize}[noitemsep,leftmargin=*]
\item \textsc{Perplexity:} 
We compute the length normalized perplexity of each answer $a^*_i$ and prefix $p$ as used in Section~\ref{subsec:single_answer}.
Then, we sort the answers in ascending order of these perplexities, resulting in `known' answers placed earlier.

    \item \textsc{Greedy:} 
We arrange the gold answers by performing a beam search decoding in a greedy manner, constrained to permissible tokens. 
There will be two loops, outer loop for selecting the first token of the generated answer, and inner loop for completing the chosen first token.

Figure \ref{alg:grd} presents the pesudocode, which we explain below. Let's denote $a^*_i$ as a sequence of tokens $(w_{i_1}, w_{i_2}, ..., w_{i_{|a^*_i|}})$ for the $i$-th answer. At each decoding step $t$, a set of permissible tokens $I_t$ is constructed. Initially, $I_1=\{w_{1_1}, w_{2_1}, ..., w_{n_1}\}$, a set of the first token for each potential answer. We choose a token from this set that has the highest likelihood given the prompt, i.e. $o_1=\text{argmax}_{w\in I_1} P(w|p)$. Then, we update the prefix $p\leftarrow [p;o_1]$. This initiates the inner loop, setting $I_2=\{w_{i_2} |  w_{i_1} = o_1 \}$ as a set of second token of answers who starts with the selected first token. This continues until one of the answers $a_{k_1}$ is fully generated. Afterwards, we come back to the outer loop, and the initial set of permissible tokens is set to be $I_1=\{w_{1_1}, w_{2_1}, ..., w_{n_1}\}\setminus\{w_{{t_1}}\}$ excluding $a^*_{k_1}$ which has been already generated. This process continues until all answers has been generated, with a time complexity of O($n|a^*_i|$).

\end{itemize}
\section{Results for Answer Ordering Strategies}
\label{sec:ordering_results}

Having introduced strategies for ordering answers for in-context examples, we study how this impacts the generation of answers with Llama2 and OPT. We first evaluate whether the generated answers mimic the ordering of answers in in-context examples. Then, we evaluate whether the ordering impacts the size and the accuracy of predicted answer set. We also report whether two model's parametric knowledges are in sync, meaning, if one model knows about one fact, does the other model likely to know the same fact? We overall observe such patterns, particularly for QUEST dataset. 

\subsection{Does the predicted answer set follow the ordering of in-context answer set?}
\label{sec5_1}

Throughout in-context learning, the model is expected to learn the pattern shown in the demonstrations~\citep{min2022rethinking}.
We assess the generated answers to observe if the model has followed the particular ordering shown in in-context examples.

\begin{figure}[tbp]
    \centering
    \includegraphics[width=0.48\textwidth]{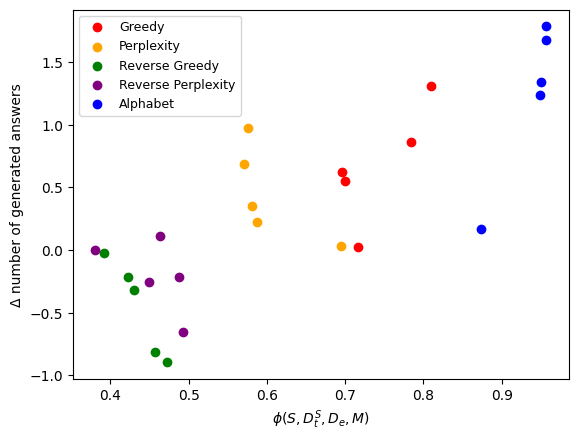}  
    \caption{
    $\phi(S, \mathcal{D}_t^{S}, \mathcal{D}_e, \mathcal{M})$ vs. the number of generated answers across three datasets, where we use Llama2 for $\mathcal{M}$. Instead of the raw number of answer set, we report the size difference compared to the answer set generated from random ordering. As $\phi$ increases, which signifies how faithfully LM follows the ordering strategy in in-context examples, the model generates more answers.}
    \label{fig:ansnum}
\end{figure}

\begin{table*}
\footnotesize
\begin{center}
\begin{tabular}{l|cccc}
\toprule
\multicolumn{1}{c|}{\textit{AmbigQA}} & $P_{EM}$ & $R_{EM}$ & $F1_{EM}$ & $F1_{F1}$ \\ \midrule
\textsc{Random} & 27.1 & 17.9 & 20.0 & 31.3 \\
\textsc{Greedy} & \blu{27.2} & \blu{\textbf{18.5}} & \blu{\textbf{20.5}} & \blu{31.7} \\
\textsc{Perplexity} & \blu{\textbf{27.4}} & \blu{18.4} & \blu{\textbf{20.5}} & \blu{\textbf{31.8}} \\
\textsc{Reverse Greedy} & 27.1 & \red{17.8} & \blu{20.1} & \blu{31.5}  \\
\textsc{Reverse Perplexity} & \blu{27.3} & 17.9 & \blu{20.2} & \blu{\textbf{31.8}}  \\
\textsc{Alphabet} & \red{26.7} & \blu{18.2} & \blu{20.3} & \red{31.2} \\  \midrule
\multicolumn{1}{c|}{\textit{QAMPARI}} & $P_{EM}$ & $R_{EM}$ & $F1_{EM}$ & $F1_{F1}$  \\ \midrule
\textsc{Random} & 26.3\: / 25.2\: & 11.7\: / 10.9\: & 13.8\: / 12.9\: & 25.3\: / 22.4\: \\
\textsc{Greedy} & \blu{26.4}\: / \blu{25.7}\: & \blu{12.2}\: / \blu{\textbf{11.9}}\star & \blu{14.2}\: / \blu{\textbf{14.0}}\star & \blu{25.6}\: / \blu{22.6}\: \\
\textsc{Perplexity} & \blu{26.7}\: / \blu{26.4}\star & \blu{12.4}\star / \blu{11.6}\star & \blu{\textbf{14.6}}\star / \blu{13.9}\star & \blu{\textbf{25.8}}\: / \blu{\textbf{22.9}}\:  \\ 
\textsc{Reverse Greedy} & \blu{26.5}\: / \blu{25.8}\: & \red{11.6}\: / \red{10.1}\star & \blu{13.9}\: / \red{12.4}\: & \red{25.1}\: / \red{21.8}\:  \\
\textsc{Reverse Perplexity} & \blu{\textbf{27.0}}\: / \blu{\textbf{26.7}}\star & 11.7\:  / \blu{11.0}\: & \blu{14.0}\: / \blu{13.3}\: & \red{25.2}\: / \blu{22.5}\:  \\
\textsc{Alphabet} & \red{24.5}\star / \red{23.5}\star & \blu{\textbf{12.7}}\star / \blu{11.8}\star & \blu{14.3}\: / \blu{13.6}\: & \red{24.7}\: / \blu{22.6}\:  \\\midrule
\multicolumn{1}{c|}{\textit{QUEST}} & $P_{EM}$ & $R_{EM}$ & $F1_{EM}$ & $F1_{F1}$ \\ \midrule
\textsc{Random} & 23.9\: / \textbf{24.8}\: & 17.9\: / 19.7\: & 18.3\: / 19.9\: & 27.2\: / 27.8\: \\
\textsc{Greedy} & \red{23.8}\: / \textbf{24.8}\: & \blu{\textbf{19.6}}\star / \blu{\textbf{20.8}}\star & \blu{\textbf{19.5}}\star / \blu{\textbf{20.6}}\star & \blu{\textbf{28.6}}\star / \blu{\textbf{28.4}}\star \\
\textsc{Perplexity} & \blu{\textbf{24.3}}\: / \textbf{24.8}\: & \blu{19.3}\star / \blu{\textbf{20.8}}\star & \blu{19.4}\: / \blu{\textbf{20.6}}\star & \blu{28.0}\: / \blu{\textbf{28.4}}\star  \\
\textsc{Reverse Greedy} & \red{22.9}\: / \red{24.5}\: & \red{17.0}\: / \red{18.4}\star  & \red{17.4}\: / \red{18.8}\star  & \red{26.3}\: / \red{26.5}\star  \\
\textsc{Reverse Perplexity} & \red{23.7}\: / \red{24.5}\:  & \red{17.3}\: / \red{19.4}\: & \red{17.7}\: / \red{19.4}\: & \red{26.4}\: / \red{27.1}\star  \\
\textsc{Alphabet} & \red{20.5}\star / \red{23.8}\star  & \red{17.6}\: / \blu{20.4}\star  & \red{17.0}\: / \blu{20.0}\: & \red{25.0}\star / \red{27.0}\star \\
\bottomrule
\end{tabular}
\caption{QA performance for answer ordering strategies on Llama2 (13B) model.
$P_{EM}$ and $R_{EM}$ are precision and recall for calculating $F1_{EM}$. We present development set performance and then test set performance in each cell.
Blue color indicates improved performance compared to Random and red indicates the opposite. We put $^*$ on scores that are significantly different from that of Random ordering.
}
\vspace{-8pt}
\label{tab:answer_ordering_llama}
\end{center}
\end{table*}

\paragraph{Metric} We introduce a metric $\phi(S, \mathcal{D}_t^{S^t}, \mathcal{D}_e, \mathcal{M})$. This measures how much LM $\mathcal{M}$ follows the answer ordering strategy $S$ on evaluation dataset $\mathcal{D}_e$ when using in-context examples from training dataset $\mathcal{D}_t$ whose answered are ordered according to $S^t$.\footnote{We assume retrieving five most similar in-context examples for each evaluation example throughout this study.} When $S$ matches $S^t$, this metric will measure how much predicted outputs mimic the answer ordering strategy of in-context examples. 

Let's denote $\hat{a}_i = \{a_{i_1}, a_{i_2}, ..., a_{i_m}\}$ be the list of predicted $m$ answers for the $i$-th example of an evaluation dataset $\mathcal{D}_e$, following its generation order from model $M$. 
We reorder the predicted answers from $\hat{a}_i$ with respect to $S$ and denote $f(a_{ij})$ to be the index of $a_{ij}$ in the newly ordered set.

For each consecutive answer pair in $\hat{a}_i$, we evaluate whether their order is preserved after reordering. Then we count the number of consecutive answer pairs that have preserved the ordering, which is $P_i = \sum^{m-1}_{j=1}\mathbbm{1}(f(a_{ij})<f(a_{(i(j+1)})$.
Similarly, $N_i = \sum^{m-1}_{j=1}\mathbbm{1}(f(a_{ij})>f(a_{(i(j+1)})$ represents the number of pairs that violate the ordering. Then, we compute micro average over $\mathcal{D}_e$. 
\begin{equation*}
    \phi(S, \mathcal{D}_t^{S^t}, \mathcal{D}_e, \mathcal{M}) =  \frac{100 \cdot \sum_{i \in D_{e}} P_i}{\sum_{i \in D_{e}} (P_i + N_i)}
\end{equation*}

\paragraph{Results}
Table \ref{tab:learning} presents the results for Llama2 model, and we provide the results for OPT model in Table~\ref{tab:learning_opt} in the appendix. For each $\phi(S, \mathcal{D}_t^{S}, \mathcal{D}_e, \mathcal{M})$, we also report $\phi(S, \mathcal{D}_t^{S_\text{random}}, \mathcal{D}_e, \mathcal{M})$ as a control. We found that in every cell (except for one cell in Table \ref{tab:learning_opt}), the first number is higher than the second number, suggesting that the model follows the answer ordering pattern presented in the in-context examples. We found this is particularly true for \textsc{Alphabet} ordering, which is probably the easiest pattern to learn. 

We further observe that the model is decoding answers such that it will present \textbf{confident} answer first (following the orders of \textsc{Greedy} and \textsc{Perplexity}), even when answers in in-context example is randomly ordered. Even after introducing consistent ordering (presenting less confident answer first), the model shows propensity to present confident answer first (values for \textsc{Reverse Greedy} and \textsc{Reverse Perplexity} are below chance (50) consistently). 

\subsection{Does ordering impact the number of generated answers?}
\label{sec5_2}
Unlike in simpler QA tasks where there is exactly one gold answer, models have to decide how many answers to generate. Would consistent ordering of answers allow the model to generate more answers? 

We report the number of generated answers for each ordering strategy for Llama2 model in Figure \ref{fig:ansnum}. We find that generation order impacts the number of generated answer, with \textsc{Alphabet} ordering substantially increasing the number of generated answers the most. The results further suggest that an ordering pattern that is easier for the model to learn can prompt LM to generate more answers. We report the results for OPT model in Figure \ref{fig:ansnum_opt} which shows the same trends.

\begin{figure*}[tbp]
    \centering
    \includegraphics[width=\textwidth]{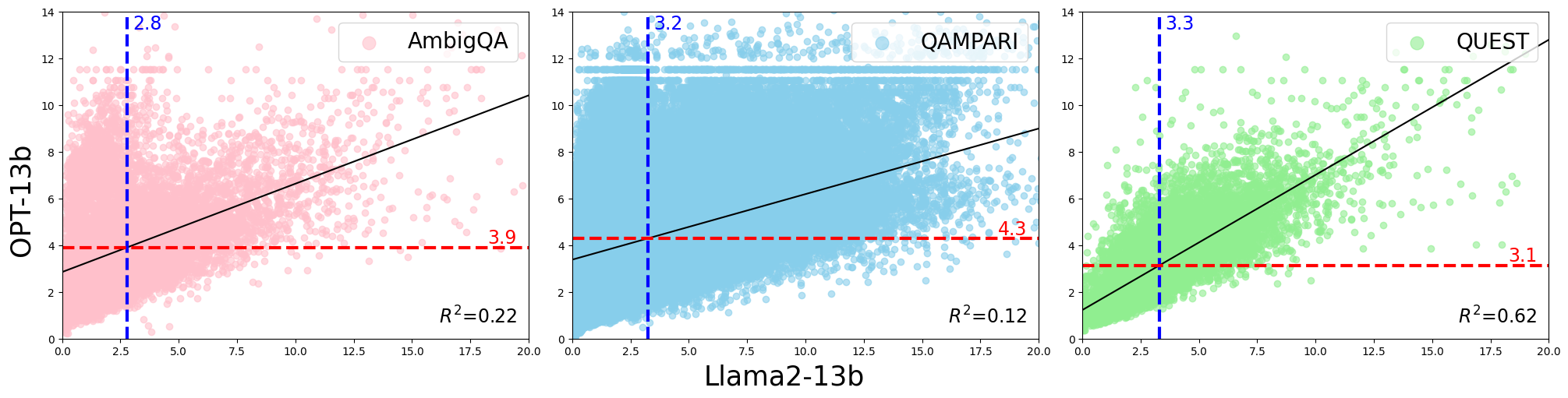}    
    \caption{Plots of log answer perplexities from Llama2-13b (x-axis) and OPT-13b (y-axis). Horizontal and vertical lines indicate the mean value of log perplexities with respect to each LM. In all datasets, Llama2 outperforms OPT in its parametric knowledge, and the answers mostly report higher perplexity with OPT compared to Llama2. } 
    \label{fig:viz_lxo}
\end{figure*}

\subsection{Does the ordering impact the QA performance?}
\label{sec5_3}
Lastly, we examine the end task (QA) performance of different answer ordering strategies. Table~\ref{tab:answer_ordering_llama} presents the results on Llama2 model. Overall, we see that answer ordering does not bring large impact in final performance, but notice consistent patterns. Presenting more confident answers first (\textsc{greedy} and \textsc{perplexity}) yielded better results than their \textsc{reverse} counterparts. \textsc{Greedy} and \textsc{Perplexity} show gains mostly in recall, leading to increase in both $F1_{EM}$ and $F1_{F1}$. Arbitrary, yet consistent ordering such as \textsc{Alphabet} does not improve model performance, sometimes rather leading to lower performance. The trend holds for AmbigQA though not statistically significant. This might be caused by smaller average answer set size compared to that of other datasets (2-3 vs. 10+ answers). We suggest ordering `known' answer first in in-context examples to improve model performance. 

For OPT model, we observe \textsc{Greedy} and \textsc{Perplexity} show improved performance through gains in recall for QUEST dataset but the results are mostly random on other datasets (Table \ref{tab:opt} in the appendix). We plot the perplexity of individual answer in train examples with respect to two models in Figure \ref{fig:viz_lxo}. Overall, we find that Llama2 contains more factual knowledge than OPT, resulting in higher end task performance. Two models exhibit similar knowledge for QUEST as they strongly correlate, however OPT shows a wider range of perplexities on other datasets, especially for answers that have low perplexity on Llama2. 
We hypothesize carefully ordering between answers will bring significant changes in end task performance \textbf{only} when model exhibits sufficient parametric knowledge of subset of answers. When the model is not familiar enough with the gold answers in in-context examples, knowledge-aware answer ordering might have limited effectiveness.
\subsection{Transfer to other base LMs}
\label{sec:gpt}

So far we have measured the parametric knowledge on an language model and then use the same model for in-context prompting. 
In this section, we experiment using in-context example set constructed with parametric knowledge of one language model (Llama2), see how it impacts the generation of another language model (GPT-3.5). While different LMs have different pre-training data, the relative parametric knowledge might be similar for different LMs (e.g., famous entity to one LM remains famous for another LM). This also allows us to experiment with propriety black-box LM API easily, whose prediction probability is not always available. We observe similar patterns as in the original experiments (Table~\ref{tab:gpt}), but the effect size is much smaller and not significant, potentially because of the difference in parametric knowledge between two models. 

\begin{table*}[tbp]
\footnotesize
\begin{center}
\begin{tabular}{l|cccc}
\toprule
\multicolumn{1}{c|}{\textit{AmbigQA}} & $P_{EM}$ & $R_{EM}$ & $F1_{EM}$ & $F1_{F1}$  \\ \midrule
\textsc{Random} & 28.2 & 22.1\:\: & 23.1 & 35.7 \\
\textsc{Perplexity} & \blu{28.8} & \blu{\textbf{23.1}}\star & \blu{\textbf{23.9}} & \blu{\textbf{36.5}} \\
\textsc{Reverse Perplexity} & \blu{\textbf{29.0}} & \blu{22.3}\:\: & \blu{23.5} & \red{35.3} \\
\textsc{alphabet} & \blu{28.4} & \blu{22.5}\:\: & \blu{23.5} & \blu{35.8} \\ \midrule
\multicolumn{1}{c|}{\textit{QAMPARI}} & $P_{EM}$ & $R_{EM}$ & $F1_{EM}$ & $F1_{F1}$  \\ \midrule
\textsc{Random} & 23.4 / \textbf{23.2} & 18.7\: / 18.5 & 18.4 / 18.4 & 30.1 / 28.4\:\: \\
\textsc{Perplexity} & \blu{\textbf{23.9}} / \red{22.9} & \blu{\textbf{19.5}}\: / \blu{\textbf{19.1}} & \blu{\textbf{18.9}} / \blu{\textbf{18.5}} & \blu{\textbf{30.4}} / \blu{\textbf{29.1}}\:\: \\ 
\textsc{Reverse Perplexity} & \red{23.2} / \red{23.1} & \red{18.2}\: / 18.5 & \red{18.2} / \red{18.3} & \blu{30.2} / \blu{28.5}\:\: \\ 
\textsc{alphabet} & 23.4 / \red{23.0} & \red{17.3}\star / \red{17.8} & \red{17.8} / \red{18.0} & \red{29.0}\star / \red{27.5}\:\: \\ \midrule
\multicolumn{1}{c|}{\textit{QUEST}} & $P_{EM}$ & $R_{EM}$ & $F1_{EM}$ & $F1_{F1}$  \\ \midrule
\textsc{Random} & 15.0 / 16.4 & 16.7 / 17.6\: & 14.8 / 15.8\:\: & 25.5 / 26.4\:\: \\
\textsc{Perplexity} & \blu{\textbf{16.6}} / \blu{\textbf{17.0}} & \blu{\textbf{17.7}} / \blu{\textbf{18.6}}\star & \blu{\textbf{15.9}} / \blu{\textbf{16.5}}\star & \blu{\textbf{26.8}} / \blu{\textbf{26.8}}\:\: \\
\textsc{Reverse Perplexity} & \blu{16.2} / \blu{16.5} & \blu{17.5} / \blu{17.8}\:\: & \blu{15.5} / \blu{15.9}\: & \blu{26.6} / 26.4\:\: \\
\textsc{alphabet} & \blu{15.5} / \blu{\textbf{17.0}} & \red{16.2} / 17.6\:\: & \blu{14.9} / \blu{16.2}\: & \red{24.9} / \red{25.5}\star \\ 
\bottomrule
\end{tabular}
\caption{QA performance for answer ordering strategies with GPT-3.5 model. We order the answer set with respect to parametric knowledge of Llama2 and evaluate its transfer to GPT-3.5 model.}
\vspace{-8pt}
\label{tab:gpt}
\end{center}
\end{table*}

\subsection{Random In-Context Examples}
\label{sec:random}
Prior works have highlighted the importance of relevant in-context examples, such as those based on similarity \cite{liu2021makes} and diversity \cite{Levy2022DiverseDI}. Yet, many studies do not do example specific retrieval and use random examples for its simplicity. Throughout our experiments in Section \ref{sec:ordering_results}, we retrieved similar in-context examples for each evaluation example. How would our results hold if we use randomly select in-context examples? 

First, with randomly retrieved in-context examples, models still learn to follow the answer ordering strategy shown in in-context examples but substantially less than when using similar in-context examples (Table \ref{tab:learning_random} in the appendix). Second, we find that the number of generated answer is affected similarly, with using \textsc{alphabet} ordering leads to the highest number of generated answers. However, we see invariant performances on end tasks (Table \ref{tab:random_examples} in the appendix).
Carefully constructing relevant in-context examples is more meaningful than doing it for random in-context examples. This suggests that if you do not have large enough training examples to recover semantically relevant in-context examples, careful construction of prompt might not yield changes in end task performance.
\section{Related Work}
\label{related-work}

\paragraph{Multi Label Ordering} While not studied extensively under the in-context-learning setting, a recent work~\cite{Madaan2022ConditionalSG} studies set generation problem from an encoder-decoder model, showing that imposing informative ordering over the label space improves model performance. 
\paragraph{Analysis on In-context Learning}
Many prior works investigate factors that determine the performance of in-context learning ~\cite{brown2020language}, such as the composition of the pre-training dataset ~\citep{xie2022an}, size of language model ~\citep{wei2022emergent}, number of pre-training tokens ~\citep{touvron2023llama}, and specific fine-tuning strategy employed ~\citep{wei2021finetuned}. 
More closely related to ours, one line of work particularly focuses on factors related to the in-context examples, including the choice of verbalizer and templates~\cite{min2022rethinking}, order of examples~\cite{lufantastically, Pezeshkpour2023LargeLM}, and the choice of in-context examples~\cite{liu2021makes,rubin2021learning,Agrawal2022IncontextES,ye2023compositional}. 
While past work is mainly centered around classification tasks, our work studies the task of multi-answer QA, with a focus on how LM's parametric knowledge on in-context examples impact the performance. In particular, our findings suggests that answers with lower perplexity lead to more accurate answer, which is congruent with recent work that shows using lower perplexity prompts improves model perplexity in general~\cite{Ye-Durrett:explselect,Iter2023InContextDS,Gonen2022DemystifyingPI}.

\paragraph{Multi-answer QA}
Real-world questions could naturally have multiple answers when a question is ambiguous~\cite{min2020ambigqa,stelmakh-etal-2022-asqa}, when a question is evaluated under different temporal or geographical contexts~\cite{zhang-choi-2021-situatedqa}, or when a question expects a set of answers~\cite{amouyal2022qampari,malaviya2023quest}. While most prior work tackles multi-answer QA in the open-book setting by retrieving from external corpus~\cite{shao-huang-2022-answering,sun-etal-2023-answering}, we study the problem in the close-book setting, which prompts LLMs to generate the answers based on their parametric knowledge.
\section{Conclusion}
We present comprehensive studies on knowledge-aware prompt design for multi-answer QA tasks. Our findings underscore the benefits of having in-context examples that the language model is familiar with.
First, the \textsc{Halfknown} set aids the model in effectively accessing its parametric knowledge.
Second, employing knowledge-aware ordering of presenting answers in descending order of the model's knowledge enhances the overall process of answer generation.

\newpage
\section*{Limitations}
Our study mainly focuses on multi-answer QA datasets, while we have demonstrated the potential for generalization to single-answer QA and NLI tasks. The analysis can be extended to a wide range of tasks that requires various types of reasoning abilities. Also, we find that the end task performance gets less impacted when random in-context examples are used. Further studies could explore diverse in-context example retrieval methods as well as cover multiple languages.

\section*{Acknowledgment}
We thank the members of UT Austin NLP community for valuable feedback, especially Thom Lake for suggesting an ablation study and Greg Durrett for providing feedback on the draft. This work is partially funded by a gift from Home Depot. 
This work was supported by Korea Institute for Advancement of Technology (KIAT) grant funded by the Korea Government (Ministry of Education) 
(P0025681-G02P22450002201-10054408, Semiconductor-Specialized University).

\bibliography{anthology,custom}
\clearpage
\appendix

\section{Dataset Statistics}
\label{appendix:dataset}
\begin{table*}
\footnotesize
\begin{center}
\begin{tabular}{l|rr|rrr|rrr}
\toprule
& \multicolumn{2}{c|}{AmbigQA} & \multicolumn{3}{c|}{QAMPARI} & \multicolumn{3}{c}{QUEST} \\
& Train & Dev. & Train & Dev. & Test & Train & Dev. & Test \\ \midrule
\# Examples& 4,615 & 1,048 & 50,372 & 1,000 & 1,000 & 1,251 & 316 & 1,669 \\
Avg. \# of answers & 2.8 & 3.1 & 14.0 & 13.2 & 13.1 & 10.9 & 10.7 & 10.7 \\
Query length & 46.9 & 46.7 & 67.8 & 57.7 & 55.8 & 54.0 & 52.2 & 53.3 \\
Answer length & 15.9 & 14.5 & 14.4 & 17.3 & 16.6 & 17.2 & 16.7 & 17.0 \\
Answer sequence length & 45.2 & 45.4 & 200.9 & 228.5 & 217.6 & 187.0 & 179.0 & 182.4 \\
\# Unique answers & 10,684 & 2,999 & 455,469 & 12,462 & 12,464 & 10,160 & 3,050 & 12,367 \\
\bottomrule
\end{tabular}
\caption{Dataset statistics. Lengths of query, answer, and answer sequence are measured by the length of each string. \# Unique answers counts unique answers within each split. 
Duplicated questions are removed from training sets.
}
\label{tab:dataset_statistics}
\end{center}
\end{table*}
We report the dataset statistics in Table~\ref{tab:dataset_statistics}.

\section{Similarity of In-Context Examples}

\begin{figure}[htbp]
\footnotesize
    \centering

    \includegraphics[width=0.47\textwidth]{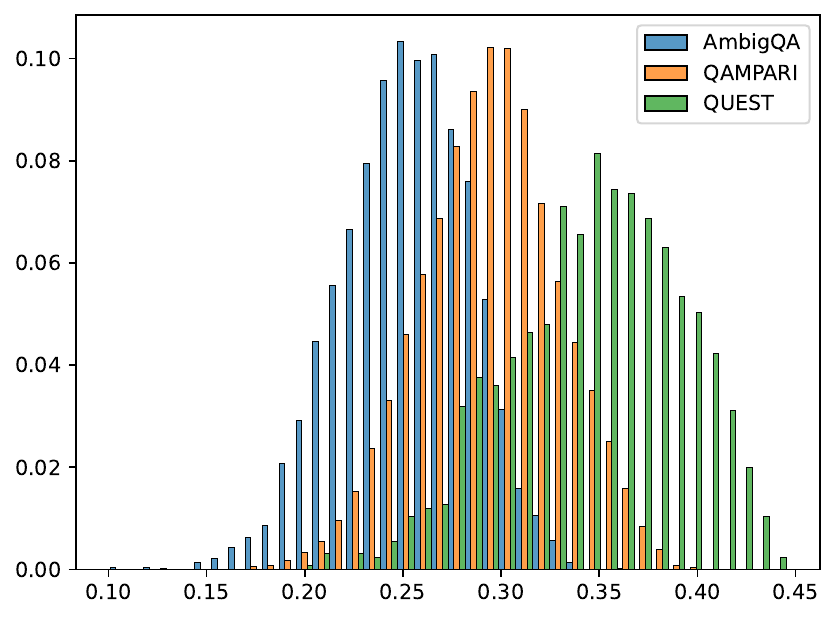} 
    \caption{Similarity distributions among in-context example candidates. 
    The x-axis denotes embedding similarity (with SimCSE~\cite{gao2021simcse} encoder) and the y-axis indicates the percentage of each bin. The median value for each dataset is 0.254, 0.295, 0.350.
    }
    \label{fig:sim_dist}
\end{figure}

We calculate the similarity score of two in-context examples using SimCSE embeddings of each query. Figure \ref{fig:sim_dist} illustrates the similarity distributions across three datasets.

\section{Experimental Details}
\subsection{Resources}
All experiments are conducted on NVIDIA A40 GPU. A single evaluation for AmbigQA and QUEST (development split) took around 20 minutes. QAMPARI (development and test split) took around 1 hours. QUEST (test split) took around 2 hours, due to its largest size.

\subsection{Statistical Testing}
We conduct paired bootstrap tests with 10000 bootstrap samples throughout our experiments (Section \ref{lab:testing}). Since we have multiple (two or four) in-context example sets for experiments in Section \ref{pkexp}, we randomly sample one in-context example set of each class (\textsc{unknown}, \textsc{halfknown}, \textsc{known}, and \textsc{random}) and conduct testing.

\section{In-Context Example Set Study}
In Table \ref{tab:answer_sets_app}, we present the results from Section \ref{sec:known_set} for QAMPARI\textsubscript{test} and QUEST\textsubscript{dev} on Llama2.

\begin{table}[htb]
\footnotesize
\begin{center}
\begin{tabular}{l|cc|cc}
\toprule
& \multicolumn{2}{c|}{QAMPARI\textsubscript{test}} & \multicolumn{2}{c}{QUEST\textsubscript{dev}} \\
& $F1_{EM}$ & $F1_{F1}$ & $F1_{EM}$ & $F1_{F1}$ \\ \midrule
Random & 10.0\:\: & 19.3\: & 4.0\: & 12.1\: \\
Unknown & 10.6\:\: & 20.2\star & 4.4\star & \textbf{13.2}\star \\ 
HalfKnown & \textbf{11.2}\star & \textbf{20.9}\star & \textbf{4.9}\star & 13.1\star \\
Known & 9.9\: & 18.6\: & 4.3\star & 12.8\star \\
\bottomrule
\end{tabular}
\caption{
Results comparing known example and unknown example. We put $^*$ on scores that are significantly different from that of Random in-context examples set, and bold the highest performing set for each metric.
}
\label{tab:answer_sets_app}
\end{center}
\end{table}

\section{Answer Ordering Strategies}
\subsection{Single Answer Study}

\begin{figure}[htbp]
    \centering
\includegraphics[width=0.5\textwidth]{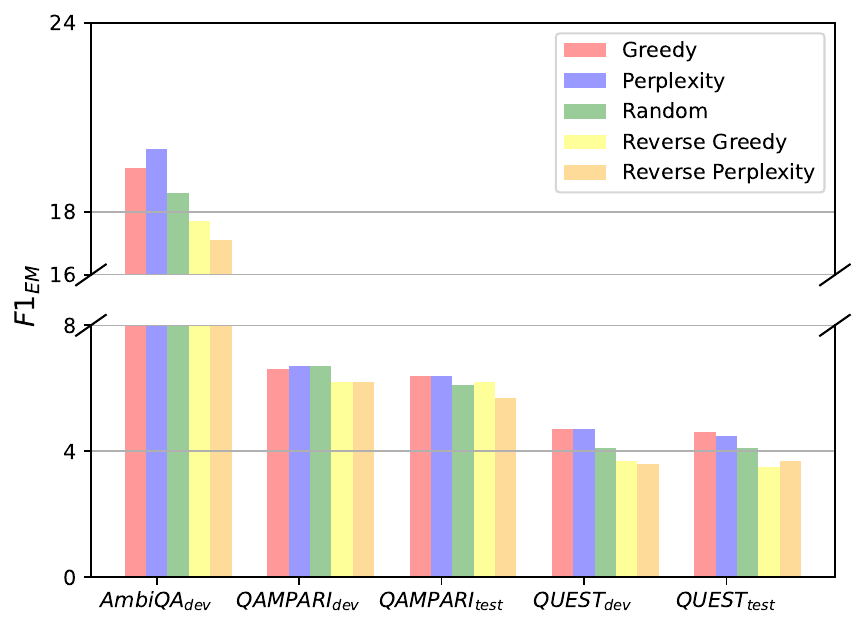}    
    \caption{Answer-level Exact Match ($F1_{EM}$) score for demonstrating only one frontmost answer of an ordering methodology on Llama2 model.}
    \label{fig:oneans_strategy}
\end{figure}

We examine the effectiveness of answer ordering strategies discussed at Section \ref{sec:ordering_strategies}. We provide only one answer at the forefront of each ordered answers in in-context examples.
Since an answer from \textsc{Greedy} and \textsc{Perplexity} is `known' to the model, they may serve as an upper bound of `known' answer, while \textsc{Reverse Greedy} and \textsc{Reverse Perplexity} may serve as a lower bound. \textsc{Random} exists somewhere between these.
The disparities among these are clear, as shown in Figure ~\ref{fig:oneans_strategy}.
The results suggest that the model is able to differentiate ordering strategies.

\subsection{Results on OPT 13B model}
\label{acc:opt}
\begin{figure}[tbp]
    \centering
    \includegraphics[width=0.48\textwidth]{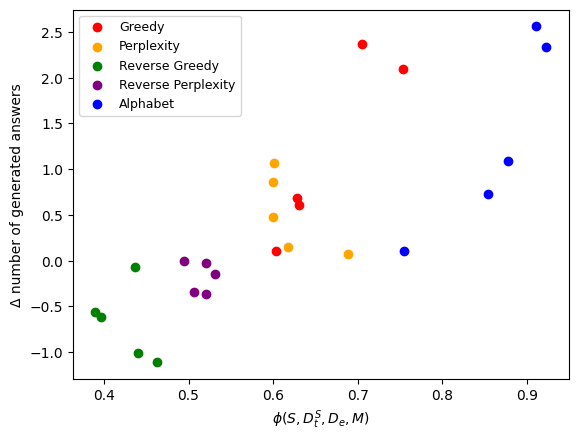} \vspace{-1.5em}
    \caption{
    $\phi(S, \mathcal{D}_t^{S}, \mathcal{D}_e, \mathcal{M})$ vs. the number of generated answers across three datasets, where we use OPT (13B) model for $\mathcal{M}$.}\vspace{-0.5em}   
    \label{fig:ansnum_opt}
\end{figure}

We present the results of experiments in Section \ref{sec:ordering_results} with OPT 13B model. 
With respect to following the ordering strategy of in-context examples (Section~\ref{sec5_1}, \ref{sec5_2}), we find that the results hold for OPT LLM model as well (Table \ref{tab:learning_opt}).
However, the end task performance results are somewhat mixed (Table \ref{tab:opt}, Figure \ref{fig:ansnum_opt}). We observe consistent results of end task performance on QUEST dataset but the results are mostly random on AmbigQA and QAMPARI dataset.

\subsection{Results on GPT-3.5 model}
GPT-3.5-turbo model tends to generate lengthy and chatty outputs such as ``There is not enough information given to answer this question". Therefore we add a short instruction as following: ``\texttt{Follow the answers pattern}". 

\begin{table*}[tbp]
\footnotesize
\begin{center}
\begin{tabular}{llccccc}
\toprule
&&&&$S$\\
& & \textsc{Greedy} & \textsc{Reverse Greedy} & \textsc{Perplexity} & \textsc{Reverse Perplexity} &\textsc{Alphabet} \\ \midrule
&AmbigQA\textsubscript{dev} & 60.3 / 58.3 & 43.7 / 42.2 & 68.8 / 58.1 &  49.5 / 41.9 & 75.5 / 50.5 \\
&QAMPARI\textsubscript{dev} & 62.8 / 52.1 & 39.0 / 39.6 & 60.0 / 55.1 & 52.1 / 44.9 & 87.8 / 52.0 \\
$\mathcal{D}_e$&QAMPARI\textsubscript{test} & 63.1 / 52.4 & 39.7 / 39.1 & 61.8 / 56.7 &  52.1 / 39.1 & 85.4 / 47.3 \\
&QUEST\textsubscript{dev} & 70.5 / 49.1 & 44.0 / 42.5 & 60.0 / 57.1 & 53.1 / 42.9 & 91.1 / 67.6  \\
&QUEST\textsubscript{test} & 75.3 / 57.5 &  46.3 / 45.5 & 60.1 / 54.0  & 50.6 / 46.0  & 92.3 / 51.6 \\\midrule
\multicolumn{2}{c}{\textbf{Average}} &  66.4 & 42.5 & 62.1 & 51.5 & {86.4} \\
\bottomrule
\end{tabular}
\caption{Percentage of generated answer ordering matching in-context examples answer ordering, where we use OPT (13B) model for $\mathcal{M}$. The table is formatted the same as Table~\ref{tab:learning}.
}
\label{tab:learning_opt}
\end{center}
\end{table*}

\begin{table*}[!ht]
\footnotesize
\begin{center}
\begin{tabular}{l|cccc}
\toprule
\multicolumn{1}{c|}{\textit{AmbigQA}} & $P_{EM}$ & $R_{EM}$ & $F1_{EM}$ & $F1_{F1}$ \\ \midrule
\textsc{Random} & 13.1 & 10.3 & 10.7 & 19.4 \\
\textsc{Greedy} & 13.1 & 10.3 & 10.7 & \textbf{\blu{19.5}} \\
\textsc{Perplexity} & \red{12.9} & \red{10.0} & \red{10.5} & \red{19.2} \\
\textsc{Reverse Greedy} & \red{12.9} & \red{9.9} & \red{10.5} & \red{19.1}\\
\textsc{Reverse Perplexity} & \blu{13.2} & \textbf{\blu{10.7}} & \textbf{\blu{11.0}} & \red{19.3} \\
\textsc{Alphabet} & \textbf{\blu{13.5}} & \blu{10.6} & \textbf{\blu{11.0}} & \red{19.3}  \\ \midrule
\multicolumn{1}{c|}{\textit{QAMPARI}} & $P_{EM}$ & $R_{EM}$ & $F1_{EM}$ & $F1_{F1}$ \\ \midrule
\textsc{Random} & 14.2\: / 15.5\: & 7.5\: / 7.2\: & 8.1\: / 8.2\: & 18.6\: / 17.1\:\\
\textsc{Greedy} & \red{14.0}\: / \red{14.9}\: & 7.5\: / \blu{7.6}\: & \red{7.9}\: / \blu{8.4}\: & 18.6\: / \textbf{\blu{17.8}}\:\\
\textsc{Perplexity} & \textbf{\blu{14.7}}\: / \blu{15.6}\: & \blu{7.8}\: / \blu{7.7}\: & \blu{8.3}\: / \blu{8.5}\: & \blu{19.0}\: / \blu{17.6}\:\\ 
\textsc{Reverse Greedy} & \blu{14.5}\: / \red{15.4}\: & \red{6.9}\star / \red{6.7}\: & \red{7.6}\: / \red{7.9}\: & \red{18.0}\star / \red{16.7}\:\\
\textsc{Reverse Perplexity} & \blu{15.6}\star / \textbf{\blu{15.9}}\: & \blu{7.6}\: / 7.2\: & \blu{8.4}\: / \blu{8.3}\: & \textbf{\blu{18.8}}\: / \red{16.9}\:\\
\textsc{Alphabet} & \blu{14.4}\: / \red{15.0}\: & \textbf{\blu{8.1}}\star / \textbf{\blu{7.9}}\star & \textbf{\blu{8.5}}\: / \textbf{\blu{8.9}}\star & \blu{18.7}\: / \blu{17.4}\:\\ \midrule
\multicolumn{1}{c|}{\textit{QUEST}} & $P_{EM}$ & $R_{EM}$ & $F1_{EM}$ & $F1_{F1}$ \\ \midrule
\textsc{Random} & 14.6\: / 18.4\: & 11.6\: / 16.1\: & 12.0\: / 15.6\: & 21.3\: / 23.8\: \\
\textsc{Greedy} & \blu{15.7}\: / \blu{\textbf{18.6}}\: & \blu{\textbf{16.6}}\star / \blu{\textbf{18.0}}\star & \blu{\textbf{\blu14.9}}\star / \blu{\textbf{17.0}}\star & \blu{\textbf{23.7}}\star / \blu{\textbf{25.2}}\star \\
\textsc{Perplexity} & \blu{16.1}\: / \red{18.3}\: & \blu{14.8}\star / \blu{17.0}\star & \blu{13.9}\star / \blu{16.2}\: & \blu{22.6}\: / \blu{24.5}\:\\
\textsc{Reverse Greedy} & \red{14.5}\: / \red{17.4}\star & \red{10.7}\: / \red{13.8}\star & \red{10.2}\: / \red{13.8}\star & \red{19.6}\: / \red{22.1}\star\\
\textsc{Reverse Perplexity} & \blu{15.0}\: / \red{17.9}\: & \blu{14.3}\star / \red{15.4}\: & \blu{13.2}\: / \red{15.1}\: & \blu{22.4}\: / \red{23.5}\:\\
\textsc{Alphabet} & \blu{\textbf{16.3}}\star / \red{17.6}\star & \blu{15.9}\star / \blu{17.3}\star & \blu{14.7}\star / \blu{16.3}\star & \blu{23.0}\star / \blu{24.1}\star\\
\bottomrule
\end{tabular}
\caption{QA performance for answer ordering strategies with OPT (13B) model. The table is formatted the same as Table~\ref{tab:answer_ordering_llama}.
}
\label{tab:opt}
\end{center}
\end{table*} 

\begin{table*}[tbp]
\footnotesize
\begin{center}
\begin{tabular}{llccccc}
\toprule
&&&&$S$\\
& & \textsc{Greedy} & \textsc{Reverse Greedy} & \textsc{Perplexity} & \textsc{Reverse Perplexity} &\textsc{Alphabet} \\ \midrule
&AmbigQA\textsubscript{dev} & 69.6 / 68.9 & 33.7 / 32.8 & 70.2 / 70.5 & 68.9 / 29.5 & 83.6 / 62.5  \\
&QAMPARI\textsubscript{dev} & 63.2 / 59.8 & 40.7 / 40.3 & 57.0 / 57.3 & 57.3 / 42.7 & 92.6 / 65.9 \\
$\mathcal{D}_e$&QAMPARI\textsubscript{test} & 61.2 / 61.4 & 43.5 / 43.2 & 57.5 / 56.4 & 57.5 / 43.6 & 92.7 / 60.7 \\
&QUEST\textsubscript{dev} & 55.4 / 52.6 & 39.3 / 40.2 & 59.1 / 57.4 & 57.1 / 42.6 & 88.5 / 59.7 \\
&QUEST\textsubscript{test} & 56.8 / 54.0  & 38.9 / 40.1 & 56.9 / 56.4 & 56.4 / 43.6 & 86.7 / 60.9 \\\midrule
\multicolumn{2}{c}{\textbf{Average}} &  61.2 & 39.2 & 60.1 & 59.4 & 88.8 \\
\bottomrule
\end{tabular}
\caption{
Percentage of generated answer ordering matching in-context examples answer ordering, where we employ {random} in-context examples instead of most similar examples. The table is formatted the same as Table~\ref{tab:learning}.
}
\label{tab:learning_random}
\end{center}
\end{table*}

\begin{table*}[!ht]
\footnotesize
\begin{center}
\begin{tabular}{l|ccc|ccc|ccc}
\toprule
& \multicolumn{3}{c|}{\textit{AmbigQA\textsubscript{dev}}} & \multicolumn{3}{c|}{\textit{QAMPARI\textsubscript{dev}}} & \multicolumn{3}{c}{\textit{QAMPARI\textsubscript{test}}} \\
& $F1_{EM}$ & $F1_{F1}$ & \# ans & $F1_{EM}$ & $F1_{F1}$ & \# ans & $F1_{EM}$ & $F1_{F1}$ & \# ans \\ \midrule
\textsc{Random} & 17.8 & \textbf{28.7} & 2.07 & \textbf{9.8} & 20.2 & 3.77 & \textbf{10.0} & \textbf{19.1} & 3.74 \\
\textsc{Greedy} & 17.4 & 27.8 & 2.12  & 9.6 & 19.9 & 4.42 & 9.3 & 17.7 & 4.43  \\
\textsc{Perplexity} & \textbf{17.9} & 28.3 & 2.11 & 9.7 & 20.0 & 3.99 & 9.7 & 18.6 & 4.03 \\
\textsc{Reverse greedy} & 17.6 & 28.3 & 2.08 & \textbf{9.8} & \textbf{20.4} & 3.82 & 9.6 & 18.5 & 3.61 \\
\textsc{Reverse perplexity} & \textbf{17.9} & 28.4 & 2.11 & 9.3 & 19.7 & 3.83 & 9.6 & 18.4 & 3.81 \\
\textsc{Alphabet} & \textbf{17.9} & 28.5 & \textbf{2.22} & \textbf{9.8} & 19.8 & \textbf{5.48} & 9.6 & 17.5 & \textbf{5.41} \\
\bottomrule
\end{tabular}

\bigskip

\begin{tabular}{l|ccc|ccc}
\toprule
& \multicolumn{3}{c|}{\textit{QUEST\textsubscript{dev}}} & \multicolumn{3}{c}{\textit{QUEST\textsubscript{test}}} \\
& $F1_{EM}$ & $F1_{F1}$ & \# ans & $F1_{EM}$ & $F1_{F1}$ & \# ans \\ \midrule
\textsc{Random} & 4.4 & 12.9 & 3.42 & 3.5 & 11.2 & 3.41 \\
\textsc{Greedy} & \textbf{4.7} & 12.5 & 4.51 & 3.4 & 10.9 & 4.49 \\
\textsc{Perplexity} & \textbf{4.7} & \textbf{13.0} & 3.60 & 3.4 & 11.1 & 3.62 \\
\textsc{Reverse Greedy} & 4.0 & 12.5 & 3.51 & 3.3 & 11.1 & 3.11 \\
\textsc{Reverse Perplexity} & 4.6 & 12.6 & 3.09 & \textbf{3.6} & \textbf{11.4} & 3.28 \\
\textsc{Alphabet} & 4.5 & 11.2 & \textbf{5.84} & 3.0 & 9.4 & \textbf{5.99} \\
\bottomrule
\end{tabular}
\caption{
QA performance for answer ordering strategies with random in-context examples. We bold the highest performing set for each metric.
}
\label{tab:random_examples}
\end{center}
\end{table*}

\section{Prompts}
\label{prompt_example}

Throughout Table \ref{tab:ex_ambiqa} to Table \ref{tab:ex_snli}, we present the prompts used in our experiments. 

\def\sn{\textbackslash n}

\begin{table*}[tbp]
\footnotesize
\begin{center}
\begin{tabular}{l|p{13cm}}
\toprule
Question & Who is the current chairman of african union commission? \\ \midrule
Gold Answers & Jean Ping, Moussa Faki, Nkosazana Clarice Dlamini-Zuma \\ \midrule
Prompt & Question: Who is the chairman of the federal reserve?\sn Answers: Alan Greenspan | Ben Bernanke | Janet Yellen\sn\sn Question: Who is the president of south africa now?\sn Answers: Thabo Mvuyelwa Mbeki | Kgalema Petrus Motlanthe | JZ\sn\sn Question: Who is the present chairperson of national human rights commission in india?\sn Answers: Justice K. G. Balakrishnan | H. L. Dattu | Cyriac Joseph\sn\sn Question: Who appoints the chairman of the finance commission?\sn Answers: the President | Pranab Mukherjee | Ram Nath Kovind | Pratibha Devisingh Patil\sn\sn Question: Who is the chairman of national commission for woman of india?\sn Answers: Lalitha Kumaramangalam | Mamta Sharma | Girija Vyas\sn\sn Question: Who is the current chairman of african union commission?\sn Answers:\\ \midrule
Output & Jean Ping | Nkosazana Dlamini-Zuma | Moussa Faki Mahamat\sn\\
\bottomrule
\end{tabular}
\caption{Prompt example of AmbigQA}
\label{tab:ex_ambiqa}
\end{center}
\end{table*}

\begin{table*}[tbp]
\footnotesize
\begin{center}
\begin{tabular}{l|p{13cm}}
\toprule
Question & What movies did Scott Z. Burns screenwrite? \\ \midrule
Gold Answers & Contagion, No Time to Die, Pu-239, Side Effects, The Bourne Ultimatum, The Informant!, The Laundromat, The Mercy, The Report \\ \midrule
Prompt & Question: Which film has Edward Burns as a member of its cast and had Edward Burns as screenwriter?\sn Answers: Ash Wednesday | Purple Violets | She's the One | Sidewalks of New York | The Brothers McMullen | The Groomsmen | Newlyweds | Looking for Kitty | No Looking Back\sn\sn Question: Scott Ziehl was a director for what genre of film?\sn Answers: action film | crime thriller | horror film | monster film | drama\sn\sn Question: What are the publication dates of film that  had Scott Z. Burns as screenwriter?\sn Answers: 2007 | 2006 | 2009 | 2013 | 2014 | 2018 | 2019 | 2020\sn\sn Question: Who are the cast members of film that  had Scott Z. Burns as screenwriter?\sn Answers: Christoph Waltz | Patton Oswalt | Marsha Stephanie Blake | Jocko Sims | David Costabile | Dan Fredenburgh | Ted Levine | Ken Stott | Steven Berkoff\sn\sn Question: What are the genres of film that  had Scott Z. Burns as screenwriter?\sn Answers: action film | thriller film | comedy film | spy film | drama\sn\sn Question: What movies did Scott Z. Burns screenwrite?\sn Answers:\\ \midrule
Output & Contagion | Side Effects | The Bourne Ultimatum | The Bourne Supremacy | The Bourne Identity\sn\\
\bottomrule
\end{tabular}
\caption{Prompt example of QAMPARI}
\label{tab:ex_qampari}
\end{center}
\end{table*}

\begin{table*}[tbp]
\footnotesize
\begin{center}
\begin{tabular}{l|p{13cm}}
\toprule
Question & Mary Stewart novels \\ \midrule
Gold Answers & A Walk in Wolf Wood, Airs Above the Ground, My Brother Michael, The Crystal Cave, The Hollow Hills, The Ivy Tree, The Last Enchantment, The Prince and the Pilgrim, The Wicked Day, The Wind Off the Small Isles, This Rough Magic, Thornyhold, Touch Not the Cat, Wildfire at Midnight \\ \midrule
Prompt & Question: 1740s fantasy novels, 1747 books, or novels by Denis Diderot\sn Answers: Niels Klim's Underground Travels | The Sofa: A Moral Tale | Jacques the Fatalist | Rameau's Nephew | La Religieuse | The Art of Cookery Made Plain and Easy | Letters from a Peruvian Woman | Madame de La Carli\textbackslash u00e8re | The Indiscreet Jewels | The Skeptic's Walk | Zadig | Mystification\sn\sn Question: Sarah Weeks novels\sn Answers: So B. It | Jumping the Scratch\sn\sn Question: 1858 British novels, 1872 fantasy novels, or Novels by George MacDonald\sn Answers: At the Back of the North Wind | Alec Forbes of Howglen | The Princess and the Goblin | David Elginbrod | The Day Boy and the Night Girl | Ranald Bannerman's Boyhood | The Princess and Curdie | Through the Looking-Glass | Phantastes | My Lady Ludlow | Sir Gibbie | Mary Marston | Doctor Thorne | Lilith | Poor Dionis\sn\sn Question: Mark Clapham novels\sn Answers: The Taking of Planet 5 | Twilight of the Gods | The Book of the War | Beige Planet Mars | Hope\sn\sn Question: Marlon James Novels\sn Answers: A Brief History of Seven Killings | John Crow's Devil | The Book of Night Women\sn\sn Question: Mary Stewart novels\sn Answers:\\ \midrule
Output & The Crystal Cave | The Hollow Hills | The Last Enchantment | The Wicked Day\sn\\
\bottomrule
\end{tabular}
\caption{Prompt example of QUEST}
\label{tab:ex_quest}
\end{center}
\end{table*}

\begin{table*}[tbp]
\footnotesize
\begin{center}
\begin{tabular}{p{2.5cm}|p{13cm}}
\toprule
Question & Every day, Wendi feeds each of her chickens three cups of mixed chicken feed, containing seeds, mealworms and vegetables to help keep them healthy.  She gives the chickens their feed in three separate meals. In the morning, she gives her flock of chickens 15 cups of feed.  In the afternoon, she gives her chickens another 25 cups of feed.  How many cups of feed does she need to give her chickens in the final meal of the day if the size of Wendi's flock is 20 chickens? \\ \midrule
Logical reasoning and Answer & If each chicken eats 3 cups of feed per day, then for 20 chickens they would need 3*20=<<3*20=60>>60 cups of feed per day.\sn If she feeds the flock 15 cups of feed in the morning, and 25 cups in the afternoon, then the final meal would require 60-15-25=<<60-15-25=20>>20 cups of chicken feed.\sn \#\#\#\# 20 \\ \midrule
Prompt & Question: Mabel lives 4500 steps directly east of Lake High school. Helen lives 3/4 the number of steps that Mabel lives, directly west of the school. What's the total number of steps Mabel will walk to visit Helen so that they can do their assignments together?\sn Answer: Helen lives 3/4 * 4500 = <<3/4*4500=3375>>3375 steps directly west of Lake High. To reach Helen, Mabel would have to walk to 4500 + 3375 = <<4500+3375=7875>>7875 steps. \#\#\#\# 7875\sn\sn Question: Mark is 7 years older than Amy, who is 15. How old will Mark be in 5 years?\sn Answer: Mark is 15 years + 7 years = <<15+7=22>>22 years old. In 5 years, he will be 22 years + 5 years = <<22+5=27>>27 years old. \#\#\#\# 27\sn\sn Question: Steve has 2 boxes of pencils with 12 pencils in each box. He gave Matt 3 more pencils than he gave to Lauren. If Steve gave 6 pencils to Lauren, how many pencils does he have left?\sn Answer: Steve started with 2 * 12 = <<2*12=24>>24 pencils. He gave Matt 6 + 3 = <<6+3=9>>9 pencils. After giving away the pencils, Steve will have 24 \textbackslash u2013 9 \textbackslash u2013 6 = <<24-9-6=9>>9 pencils left. \#\#\#\# 9\sn\sn Question: Mandy researched 42 med schools. She applied to 1/3 of the schools she researched and got into half of the schools where she applied. How many schools was Mandy accepted to?\sn Answer: First find the number of schools Mandy applied to: 42 med schools / 3 = <<42/3=14>>14 med schools Then divide that number by 2 to find the number of schools where she was accepted: 14 med schools / 2 = <<14/2=7>>7 med schools \#\#\#\# 7\sn\sn Question: Rachel is stuffing envelopes. She has eight hours to complete the task, and there are 1,500 envelopes. In the first hour, Rachel stuffs 135 envelopes. The second hour she stuffs 141 envelopes. How many envelopes will Rachel need to stuff per hour to finish the job?\sn Answer: Rachel has 1500 - 135 - 141 envelopes = <<1500-135-141=1224>>1224 envelopes remaining to stuff. Rachel has 8 hours - 2 hours = <<8-2=6>>6 hours left to finish the task. Rachel needs to stuff 1224 envelopes / 6 hours = <<1224/6=204>>204 envelopes per hour. \#\#\#\# 204\sn\sn Question: Samson is going to another town which is 140 km away. He will use his car that uses ten liters of gasoline for a distance of 70 km. How many liters of gasoline will Samson need for a one-way trip?\sn Answer: Samson will need 140 km/70 km = <<140/70=2>>2 ten liters of gasoline for a one-way trip to a town. Therefore, he will need a total of 2 x 10 liters = <<2*10=20>>20 liters of gasoline. \#\#\#\# 20\sn\sn Question: Every day, Wendi feeds each of her chickens three cups of mixed chicken feed, containing seeds, mealworms and vegetables to help keep them healthy.  She gives the chickens their feed in three separate meals. In the morning, she gives her flock of chickens 15 cups of feed.  In the afternoon, she gives her chickens another 25 cups of feed.  How many cups of feed does she need to give her chickens in the final meal of the day if the size of Wendi's flock is 20 chickens?\sn Answer:\\ \midrule
Output & Wendi gives her chickens 15 cups of feed in the morning and 25 cups of feed in the afternoon. She needs to give her chickens another 20 cups of feed in the final meal of the day. \#\#\#\# 20\sn \\
\bottomrule
\end{tabular}
\caption{Prompt example of GSM8K}
\label{tab:ex_math}
\end{center}
\end{table*}

\begin{table*}[tbp]
\footnotesize
\begin{center}
\begin{tabular}{l|p{13cm}}
\toprule
Premise & Yet, we now are discovering that antibiotics are losing their effectiveness against illness. Disease-causing bacteria are mutating faster than we can come up with new antibiotics to fight the new variations. \\ \midrule
Hypothesis & Bacteria is winning the war against antibiotics. \\ \midrule
Answer & entail \\ \midrule
Prompt & Please identify whether the premise entails the hypothesis. The answer should be exact 'entail' or 'not entail'.\sn premise: A brain-dead teenager was removed from life-support equipment after giving birth to a premature daughter.\sn
hypothesis: Pregnant women are allowed to die after they deliver their babies.\sn
answer: not entail.\sn\sn
premise: Kuwait becomes the fourth Gulf country where women as well as men can vote in elections. Saudi Arabia holds strictly limited, male-only elections.\sn
hypothesis: Women can vote in elections in Saudi Arabia.\sn
answer: not entail.\sn\sn
premise: The first Windows DNA lab outside Microsoft was established in suburban Philadelphia in June 1998.\sn
hypothesis: Microsoft was established in June 1998.\sn
answer: not entail.\sn\sn
premise: Scientists at the Genome Institute of Singapore (GIS) have discovered the complete genetic sequence of a coronavirus isolated from a Singapore patient with SARS.\sn
hypothesis: Singapore scientists reveal that SARS virus has undergone genetic changes.\sn
answer: not entail.\sn\sn
premise: Two portraits, one by Reynolds, the other by Gainsborough, since 1990 among the most looked-after works in the Art Loss Register database, were recovered this week when they were brought into Sotheby's for evaluation. Also a lithograph by Norwegian artist Edvard Munch, stolen from an Oslo art gallery in April, has been recovered, news reports said today.\sn
hypothesis: These two portraits and a lithography by Edvard Munch were recovered using the Art Loss Register database.\sn
answer: not entail.\sn\sn
premise: The deal marks the second foray into Lithuania by Philip Morris which in April beat British American Tobacco to acquire the Klaipeda Tobacco company for Dollars 40m - the biggest western investment in the Baltic states so far.\sn
hypothesis: The acquisition of the Klaipeda Tobacco company is the biggest investment in the Baltic states.\sn
answer: not entail.\sn\sn
premise: Yet, we now are discovering that antibiotics are losing their effectiveness against illness. Disease-causing bacteria are mutating faster than we can come up with new antibiotics to fight the new variations.\sn
hypothesis: Bacteria is winning the war against antibiotics.\sn
answer: \\ \midrule
Output & entail. \\ 
\bottomrule
\end{tabular}
\caption{Prompt example of RTE}
\label{tab:ex_rte}
\end{center}
\end{table*}

\begin{table*}[tbp]
\footnotesize
\begin{center}
\begin{tabular}{l|p{13cm}}
\toprule
Premise & This church choir sings to the masses as they sing joyous songs from the book at a church. \\ \midrule
Hypothesis & A choir singing at a baseball game. \\ \midrule
Answer & no \\ \midrule
Prompt & Please identify whether the premise entails the hypothesis. The answer should be exact 'yes', 'no' or 'neutral'.\sn
premise: A man skis past another man displaying paintings in the snow.\sn
hypothesis: A man skis past his brother.\sn
answer: neutral.\sn\sn
premise: A little boy holding a yellow, plastic shovel, crouches in the sand.\sn
hypothesis: A little girl is holding a shovel and crouches in the sand.\sn
answer: no.\sn\sn
premise: A little girl wearing a yellow coat, striped scarf, pink rain boots, and carrying a white purse is walking with her Golden Retriever down an icy and puddled road.\sn
hypothesis: A little girl, wearing a yellow coat, is walking her dog down a road.\sn
answer: yes.\sn\sn
premise: Some African American young adults are playing volleyball.\sn
hypothesis: People playing baskeball.\sn
answer: no.\sn\sn
premise: A brown and black dog is laying on a shaggy rug.\sn
hypothesis: A white dog is lying on a wooden floor.\sn
answer: no.\sn\sn
premise: Two black and white dogs in a field of flowers and grass.\sn
hypothesis: Some animals are outdoors.\sn
answer: yes.\sn\sn
premise: This church choir sings to the masses as they sing joyous songs from the book at a church.\sn
hypothesis: A choir singing at a baseball game.\sn
answer: \\  \midrule
Ouput & no. \\ 
\bottomrule
\end{tabular}
\caption{Prompt example of SNLI}
\label{tab:ex_snli}
\end{center}
\end{table*}

\end{document}